\documentclass{iaesarticle}
\shorttitle{Short Title}

\usepackage{amsmath, amsfonts, amssymb, float, fancyhdr}
\usepackage[figuresright]{rotating}
\usepackage{authblk, graphicx, indentfirst, lastpage, lipsum}
\usepackage{pifont}

\makeatletter
\renewcommand{\@biblabel}[1]{[#1]\hfill}
\renewcommand\AB@authnote[1]{\textsuperscript{\normalfont\bfseries#1}}
\makeatother
\usepackage[font=normalsize=]{subfig, caption}
\usepackage{epstopdf}
\usepackage[left=3cm, right=2.5cm, top=2.5cm, bottom=2.5cm, includehead, includefoot]{geometry}
\usepackage[justification=centering]{caption}
\usepackage{titlesec}
\usepackage[table]{xcolor}
\titleformat{\section}
{\normalfont\normalsize\bfseries\uppercase}{\thesection}{1.7em}{} 
\titleformat{\subsection}
{\normalfont\normalsize\bfseries}{\thesubsection}{.95em}{}
\titleformat{\subsubsection}
{\bfseries}{\thesubsubsection}{.2em}{}
\titlespacing*{\section}{0cm}{0.7cm}{0cm}
\usepackage{enumitem}
\usepackage[numbers,sort&compress]{natbib}
\usepackage{ragged2e}
\usepackage{hyperref,graphicx}
\usepackage{multirow}
\usepackage{fleqn}

\author[1]{\bfseries Abu Rafe Md Jamil}
\author[1]{\bfseries Nayan Malakar}

\affil[1]{Department of Computer Science and Engineering, Jashore University of Science and Technology,Jashore,Bangladesh}

\title{MLPTR-CC: Multi-label Pathology Test Recommendation using Classifier Chains and SHAP}
\shorttitle{MLPTR-CC}

\begin{document}
\setcounter{page}{1}

\setlength{\parindent}{1.27cm}

\pagestyle{fancy}
\fancyhfoffset{0cm}

\journalname{Under Review}
\vol{99}
\no{1}
\months{Month}
\years{2099}
\issn{2252-8938}
\DOI{10.11591}
\pagefirst{1}
\pagelast{1x}

\maketitle

\hrule
\vspace{.1em}
\hrule
\vspace{.5em}
\noindent
\parbox[t][][s]{0.315\textwidth}{%
\textbf{Article Info}
\vspace{.5em}
\hrule
\vspace{.5em}




\vspace{.5em}
\begin{keyword} 
\vspace{.5em}

Diagnostic test\sep Multi-label classification \sep Classifier chains \sep Ensemble learning \sep SHAP

\vspace{.5em}
\end{keyword}
\vspace{\fill}
}
\parbox{0.020\textwidth}{\hspace{0.5em}}
\parbox[t][][s]{0.65\textwidth}{
\begin{abstract}
\vspace{.3em}
Diagnostic decision making often relies on a sequence of pathology tests that bridge patient symptoms and final disease diagnosis. Existing clinical decision-support systems typically focus on predicting single diseases and do not explicitly recommend sets of intermediate tests or model dependencies among them. In this paper, we formulate pathology test recommendation as a multi-label classification problem where each case is associated with multiple, interdependent tests. We propose an AI-based framework that applies classifier chains with logistic regression, decision trees, random forests, and their ensemble to capture label dependencies between tests. Experiments on an expert-curated dataset from a private pathology laboratory show that classifier-chain models outperform their independent counterparts, improving F1-score and reducing Hamming loss while maintaining high accuracy across common and rare tests. To enhance trust and transparency, we integrate SHAP-based explainable AI, providing symptom-level attributions that align with established clinical reasoning in most cases. The results demonstrate that classifier chains combined with SHAP offer an effective and interpretable approach for multi-label pathology test recommendation, with potential to support clinicians in selecting appropriate diagnostic tests at an early stage.
\end{abstract}
}
\parbox[l]{\textwidth}{%
\rule{0.275\textwidth}{0.5pt} \hspace{0.5cm} \hrulefill
\\
\emph{\textbf{Corresponding Author:}}
\vspace{.5em}\\
Abu Rafe Md Jamil\\
Department of Computer Science and Engineering, Jashore University of Science and Technology\\
Jashore-7408, Bangladesh
\\
Email: arm.jamil@just.edu.bd
}
\vspace{.5em}
\hrule
\vspace{.1em}
\hrule


\section{Introduction}
\label{}
A better treatment outcome for patients depends on accurate and timely diagnosis. However, current diagnostic workflows are often inefficient: patients must visit physicians multiple times, test ordering varies with individual expertise, and inappropriate tests may be prescribed while critical ones are omitted. This leads to delayed decision-making, prolonged treatment cycles, and increased healthcare costs. To address these issues, we propose a machine learning–based decision-support system that recommends appropriate pathology tests from patients’ symptoms, aiming to reduce diagnostic delays, standardize test ordering practices, and support physicians rather than replace them.

Most existing machine learning approaches in healthcare focus on predicting final diseases directly from symptoms, laboratory values, or medical images, without explicitly modeling the intermediate step of ordering diagnostic tests. In many cases, these models treat outputs as independent labels and minimize or bypass expert involvement in test selection. By contrast, our work targets this underexplored intermediate stage: symptom-based recommendation of diagnostic tests that can guide timely and accurate diagnosis in primary care. Currently, pathology test selection largely depends on the physician’s individual judgment, which can be subjective and inconsistent. An AI system that recommends tests from symptoms can assist physicians, reduce variability, and improve the overall efficiency and quality of care.

Although machine learning has been widely applied to disease prediction and medical image analysis, comparatively less attention has been paid to optimizing the ordering of pathology tests, especially when multiple tests must be considered simultaneously. Prior studies have employed classifiers such as decision trees and random forests for single-label disease prediction; however, they typically ignore the complexities of multi-label classification and the dependencies among recommended tests. This highlights the limitations of single-label approaches and motivates advanced multi-label methods. Building on this gap, we adopt the classifier chain (CC) technique, which explicitly encodes dependencies between test labels by chaining binary classifiers in a specified order.

In this study, we formulate pathology test recommendation as a multi-label classification problem, where each patient case may require multiple tests. To handle label dependencies, we apply the CC approach, which sequentially conditions each classifier on both symptom features and previously predicted tests. We construct a feature dataset comprising 143 symptoms and 80 pathology tests with input from senior healthcare professionals, and we evaluate several machine learning algorithms—logistic regression, decision trees, and random forests—as base learners within the CC framework. We also employ a majority-voting ensemble to combine the strengths of individual classifiers. To ensure transparency and clinical acceptability, we integrate explainable AI (XAI) using SHAP, illustrating how individual symptoms contribute to specific test recommendations. This explainability component helps clinicians understand and validate the model’s reasoning, aligning the system with medical principles of accountability and trust.

Our experimental results show that the proposed machine learning–based system can substantially improve the accuracy and efficiency of pathology test recommendations compared with independent classifiers. In particular, the CC-based logistic regression model achieves high accuracy and low Hamming loss, while the majority-voting ensemble also delivers strong performance with high accuracy and precision. These findings demonstrate the potential of multi-label classifier chains, combined with SHAP-based explanations, to automate and enhance pathology test recommendation in primary care settings. By integrating predictive performance with interpretability, the system offers a practical and ethically defensible approach to AI-driven decision support for diagnostic testing.

\section{Literature Review}\label{sec2}

Machine learning underpins many contemporary healthcare diagnostic and predictive systems. Symptom-based models have improved evidence-based decision-making by predicting diseases from patients’ presenting complaints. However, while disease classification and outcome prediction have been widely studied, the task of recommending pathology tests from symptoms remains largely unexplored. This highlights the need for modular diagnostic systems that can suggest combinations of tests and explicitly model co-dependent diagnostic outcomes through multi-label learning.

Several systems have applied classical machine learning classifiers for symptom-based disease prediction. Decision trees, random forests, and Naïve Bayes have achieved around 95\% accuracy on the Kaggle symptoms-to-disease dataset~\cite{grampurohit2020disease}, and similar models have been used for heart-disease prediction on the UCI StatLog dataset~\cite{beyene2018survey}. GUI-based tools using Naïve Bayes and decision trees~\cite{mallela2021disease}, as well as web-based systems combining Naïve Bayes, random forests, logistic regression, and KNN for disease and doctor recommendation~\cite{kumar2021disease}, have also been proposed. Other works employ neural networks~\cite{gavhane2018prediction} and ensemble learning~\cite{ramalingam2018heart} to improve diagnostic accuracy. Nonetheless, these approaches focus on single-disease or binary outcomes and do not model dependencies among multiple diagnostic tests.

Later studies have refined predictive accuracy but still treat outputs independently. Decision tree–based “AI therapy” systems~\cite{chauhan2008disease}, feature-selection–enhanced models using SVM-RFE and Gain Ratio~\cite{pahwa2017prediction}, and multi-disease prediction with KNN, SVM, and logistic regression~\cite{rahman2023predicting} all remain primarily multi-class rather than multi-label. Research targeting specific diseases, such as heart disease~\cite{ramalingam2018heart,marimuthu2018review,shah2020heart,singh2020heart}, further optimises single-label prediction. Overall, prior healthcare-oriented work predominantly relies on single-label classification paradigms.

By contrast, multi-label learning explicitly captures correlations among outputs. The classifier chain (CC) technique models each label as a binary classifier conditioned on input features and previously predicted labels~\cite{read2011classifier}, and extensions such as the Ensemble of Classifier Chains (ECC) and ML-KNN have shown strong performance in capturing label dependencies~\cite{zhang2007mlknn,read2011classifier}. Multi-label active learning has been applied to heart-disease prediction with modest F1 improvements~\cite{el2022multi}, while other studies using random forests with dynamic rough set reduction remains single-label~\cite{shahin2014data}. Despite these advances, none of these works address pathology test recommendation as an intermediate diagnostic task requiring prediction of multiple interrelated tests from a single symptom profile.

\textbf{Research Gap and Present Study}

In summary, existing research demonstrates the potential of machine learning to improve diagnostic accuracy but largely frames problems as single-label or multi-class disease prediction. Real clinical practice, however, often requires recommending multiple, interdependent pathology tests for a given set of symptoms. Prior studies do not explicitly model label correlations for test recommendation or apply multi-label methods such as CC in this context.

To address this gap, the present study introduces a multi-label classification framework based on classifier chains and ensemble learning for pathology test recommendation. The proposed system captures dependencies among tests and recommends multiple relevant tests from patient-reported symptoms, aiming to reduce diagnostic delays, avoid unnecessary testing, and support physicians through transparent, data-driven decision assistance.
\section{Methodology}\label{sec3}

\begin{figure}[ht]
\centering
\includegraphics[width=1\linewidth]{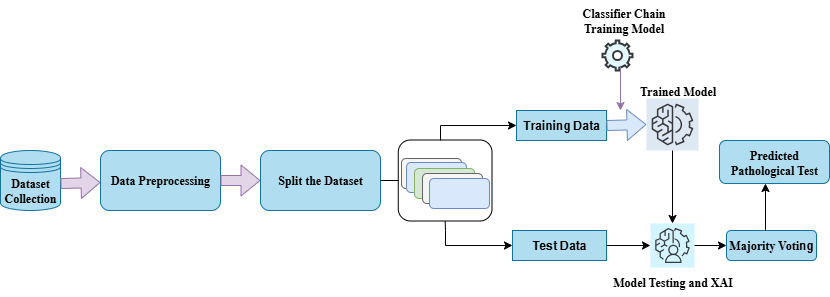}
\caption{System architecture of the proposed classifier chain–based model}
\label{fig:system_architecture}
\end{figure}

The objective of this study is to develop a machine learning–based system that can recommend suitable pathology tests based on self-reported symptoms. In routine diagnostic workflows, patients often visit healthcare facilities multiple times—first for an initial consultation, then for diagnostic tests such as blood work or imaging, and finally for follow-up review. This repetitive process delays diagnosis, increases costs and effort for patients, and places additional pressure on healthcare resources. To address these challenges, we formulate pathology test recommendation as a multi-label classification problem and adopt a classifier chain (CC) framework to predict appropriate tests before consultation with a physician. By suggesting test recommendations early in the process, the system aims to shorten time to diagnosis, make visits more efficient, and support timely clinical decision-making rather than replace clinicians.

The CC algorithm is chosen to capture inherent dependencies among tests that frequently co-occur in practice. To further improve robustness and reduce individual model bias, we combine multiple CC-based classifiers using a majority-voting ensemble strategy, yielding more stable and consistent predictions. Fig.~\ref{fig:system_architecture} illustrates the architecture of the proposed system, showing the progression from symptom inputs to multi-label test recommendations through chained classifiers and ensemble aggregation.

\subsection{Classifier Chain (CC) Technique}

We employ the classifier chain (CC) strategy to address the multi-label nature of pathology test prediction. Rather than training completely independent binary classifiers for each test label, CC builds a sequential chain of classifiers, where each classifier uses the original symptom features and the predicted outputs of all preceding classifiers as additional inputs. This structure allows the model to capture contextual dependencies among related tests, leading to more coherent and clinically plausible recommendations. For example, liver function tests and bilirubin tests are often ordered together; the CC framework can capture this dependency, whereas independent classifiers cannot.

Let $\mathbf{X} = [x_1, x_2, \ldots, x_n]$ denote the input features (symptoms) and $\mathbf{Y} = [y_1, y_2, \ldots, y_m]$ the target labels (pathology tests), where $y_j \in {0,1}$ is a binary indicator (recommended vs not recommended). Each training instance $(x_i, y_i)$ corresponds to a patient with a subset of required tests $Y_i \subseteq \mathbf{Y}$. The learning objective is to induce a function

\begin{equation}
    f : \mathbf{X} \rightarrow 2^{\mathbf{Y}}
\end{equation}
 
that maps symptom profiles to sets of recommended tests.

In the CC approach, we train a sequence of $m$ binary classifiers $(h_1, h_2, \ldots, h_m)$ such that each $h_j$ predicts $y_j$ using both the original features and the predictions of preceding classifiers:

\begin{equation}
    h_j(x) \;=\; P\!\left(y_j \,\middle|\, x, y_1, y_2, \ldots, y_{j-1}\right).
\end{equation}
This sequential dependency enables the model to learn label correlations, which are crucial in medical scenarios where tests are often prescribed dependently.

\subsubsection{Correlation-Based Ordering}

Before training, we ordered the test labels according to their pairwise correlation coefficients, computed from the binary label matrix. Labels with higher average correlation to other labels were placed earlier in the chain so that their predictions could influence subsequent classifiers. This heuristic encourages strongly connected tests to propagate their information throughout the chain, improving the consistency and interpretability of downstream predictions.

\subsubsection{Sequential Training}

The classifiers in the chain are trained sequentially. The first classifier uses only the symptom features, whereas each subsequent classifier uses the same features augmented by the predicted labels from all previous classifiers in the chain. This incremental training process allows the model to recognise cumulative relationships among related tests and loosely mimics the way physicians update diagnostic hypotheses as new test results become available.

\subsection{Multi-Class Label Adaptation}

Although the CC framework is natively binary per label, a small number of targets in our dataset had multi-class outputs (for example, graded or panel-based tests). These were handled with a one-vs-rest decomposition, recoding each multi-class target into multiple binary labels. This approach allows the CC framework to process all targets uniformly while preserving the overall multi-label formulation.

\subsection{Ensemble Learning (Majority Voting)}

To enhance robustness, we combined three CC-based models—CC with decision trees (CC-DT), CC with random forests (CC-RF), and CC with logistic regression (CC-LR)—using an ensemble majority-voting strategy at the label level. For each pathology test, the final prediction is 1 if at least two of the three base models predict 1; otherwise, the prediction is 0. This integration mitigates individual model biases and improves stability and generalisation across heterogeneous patient profiles.

\subsection{Model Selection and Base Learners}

To evaluate the robustness of the CC framework, we selected three widely used classifiers as base learners:

\begin{itemize}
\item \textbf{Decision tree (DT)} provides interpretable decision paths and can model non-linear relationships, offering transparency that aligns with clinical reasoning.
\item \textbf{Random forest (RF)} aggregates multiple decision trees to reduce variance and overfitting, yielding strong generalisation across diverse symptom patterns.
\item \textbf{Logistic regression (LR)} is a probabilistic linear classifier that serves as a stable baseline for binary decisions and is efficient to train.
\end{itemize}

All models were implemented using standard machine learning libraries, and key hyperparameters (such as tree depth, number of trees, and regularisation strength) were tuned empirically to balance precision and recall.

\begin{table}[t]
\centering
\caption{Example symptoms and corresponding pathology tests}
\label{tab:symptoms_tests}
\renewcommand{\arraystretch}{1}
\begin{tabular}{|l|l|}
\hline
\textbf{Symptom} & \textbf{Pathology tests} \\ \hline
Itching & Allergy test, liver function test, CBC \\ \hline
Skin rash & Allergy test, ANA, CBC \\ \hline
Chills & CBC, blood culture, malaria antigen test \\ \hline
Vomiting & Electrolyte panel, liver function test, CBC \\ \hline
Shivering & CBC, blood culture, malaria antigen test \\ \hline
\end{tabular}
\end{table}
\subsection{Dataset Description}

We collected data from a private pathology laboratory and constructed a custom dataset with the guidance of senior medical professionals to ensure that symptom–test mappings were clinically meaningful. The dataset comprises 223 columns, including 143 symptom attributes and 80 pathology tests (Table~\ref{tab:symptoms_tests}). Each patient record is encoded as a binary vector, where a value of 1 indicates that the patient presents a symptom or that a test was ordered, and 0 indicates absence.

To promote generalisation across diverse patient profiles, the records were randomly shuffled and split into training and test sets. In each experiment, 70\% of the data were used for training and 30\% for testing, ensuring that a wide range of symptom–test combinations appeared in both sets and reducing the risk of overfitting.

\begin{figure*}[!b]
\centering
\includegraphics[width=0.95\textwidth]{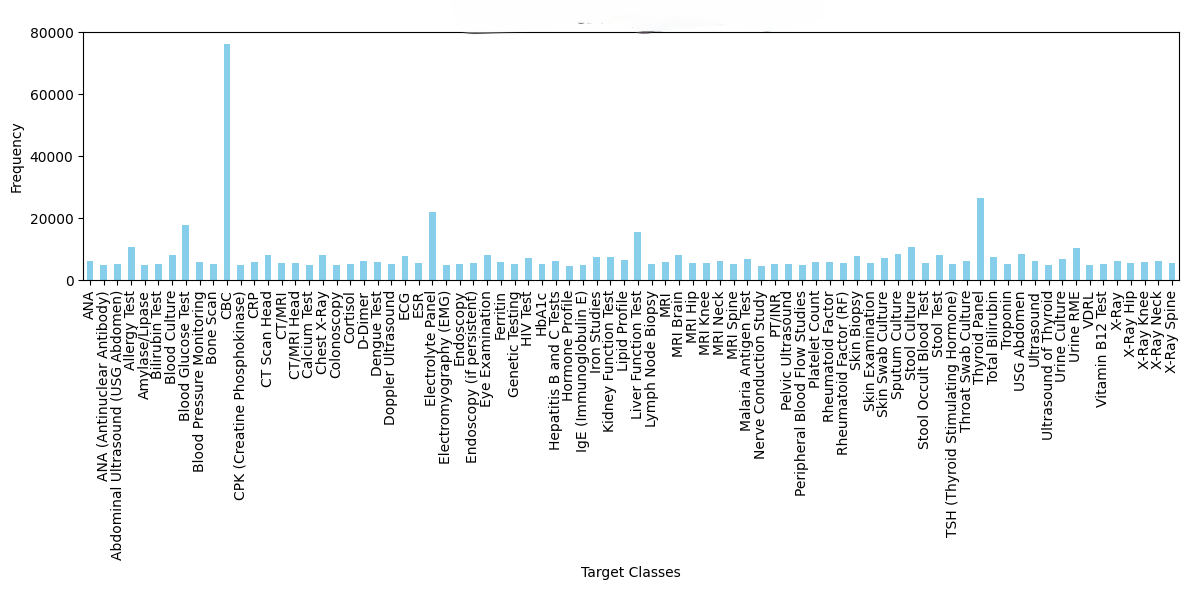}
\caption{Frequency distribution of pathology test labels}
\label{fig:Imbalanece dataset}
\end{figure*}

\subsection{Data Preprocessing}

We applied several preprocessing steps to improve data quality and ensure compatibility with the learning algorithms:

\begin{itemize}
\item \textbf{Standardisation}: For continuous features, we normalised values using the \texttt{StandardScaler} method so that features shared a comparable scale and no single feature dominated model training due to magnitude differences.
\item \textbf{Data cleaning}: Missing values were imputed with the most frequent category for the corresponding feature, preserving dataset consistency without discarding records.
\item \textbf{Class imbalance}: The dataset exhibits substantial class imbalance: some tests are ordered frequently for a wide range of symptoms, whereas others are specialised and rarely prescribed (Fig.~\ref{fig:Imbalanece dataset}). Given the multi-label structure and the desire to preserve realistic label distributions, we did not apply oversampling or downsampling. Instead, we evaluated models directly on the naturally imbalanced data and interpret results with respect to both frequent and rare tests.
\end{itemize}

\subsection{Training Procedure}

Each CC-based model was trained on 70\% of the dataset, with the remaining 30\% reserved for testing. Before training, all records were randomly shuffled in each run to expose the models to varied symptom–test patterns and reduce ordering bias. For features requiring scaling, we applied standardisation to zero mean and unit variance to ensure stable optimisation.

Within the CC framework, we first trained the initial classifier on the original symptom features. Subsequent classifiers were trained on an extended feature space that included both the symptoms and the predicted labels from earlier classifiers, thereby modelling label dependencies. This sequential process loosely emulates clinical reasoning, in which the outcome of one diagnostic test may influence the decision to order additional tests.

Hyperparameters for each base learner (DT, RF, LR) were tuned through repeated trials, and each experiment was repeated with different random seeds to assess stability. The final reported results correspond to the average performance across runs, providing a robust estimate of model behaviour on the test data.

\section{Results and Discussion}

We evaluated all models on the multi-label pathology test recommendation task using Hamming loss, sample-based accuracy, micro-averaged precision, recall, and F1-score. Each model was trained and tested on the naturally imbalanced dataset to reflect real-world diagnostic practice, where some tests are prescribed far more frequently than others. This setup allows us to assess both overall performance and robustness under realistic label distributions.

\subsection{Comparative model performance}

\begin{table}[!b]
\centering
\caption{Performance comparison of models on the multi-label test recommendation task}
\label{tab:cc_results_simple}
\renewcommand{\arraystretch}{1.15}
\begin{tabular}{|l|c|c|c|c|c|}
\hline
\textbf{Model} & \textbf{Hamming loss} & \textbf{Accuracy} & \textbf{Precision} & \textbf{Recall} & \textbf{F1-score} \\
\hline
DT & 0.8054 & 19.46\% & 0.79 & 0.57 & 0.66 \\ \hline
DT (CC)    & 0.0240 & 97.60\% & 0.82 & 0.84 & 0.83 \\ \hline
LR & 0.0563 & 45.32\% & 0.95 & 0.71 & 0.80 \\ \hline
\textbf{LR (CC)} & \textbf{0.0117} & \textbf{98.83\%} & \textbf{0.97} & \textbf{0.86} & \textbf{0.91} \\ \hline
RF & 0.7721 & 22.79\% & 0.89 & 0.59 & 0.70 \\ \hline
RF (CC)    & 0.0151 & 98.49\% & 0.92 & 0.85 & 0.89 \\ \hline
MV & 0.5958 & 40.42\% & 0.89 & 0.70 & 0.78 \\ \hline
MV (CC)    & 0.0149 & 98.53\% & 0.93 & 0.85 & 0.89 \\ \hline
\end{tabular}
\end{table}

As shown in Table~\ref{tab:cc_results_simple}, classifier-chain variants of all base models substantially improve performance over their independent counterparts. For each classifier, the CC version achieves lower Hamming loss and higher accuracy and F1-score, indicating that explicitly modelling dependencies among pathology tests leads to more accurate and contextually appropriate recommendations. Among all models, logistic regression with CC achieves the best overall performance, with the lowest Hamming loss and highest F1-score, suggesting that a probabilistic linear model augmented with label dependencies can effectively capture relationships between symptoms and tests without overfitting.

Random forest with CC and the majority-voting ensemble also obtain high performance, only slightly below that of CC-LR. Their results indicate strong generalisation and stable performance across heterogeneous symptom–test patterns. In contrast, the decision tree without CC performs poorly, underscoring the limitations of independent single-label classifiers in multi-label medical diagnostics. However, the CC version of the decision tree improves accuracy from 19.46\% to 97.60\%, demonstrating that label chaining can substantially enhance even relatively simple base learners.

To assess performance in a realistic clinical setting, we evaluated all models on the imbalanced dataset drawn from a few frequently prescribed tests (Fig.~\ref{fig:Imbalanece dataset}). Despite the skewed label distribution, the CC-based models maintain high accuracy and favourable Hamming loss, indicating that the CC framework effectively captures inter-label dependencies and improves prediction quality for both frequent and less common tests. The majority-voting ensemble further enhances robustness by aggregating predictions from multiple CC-based learners, reducing the influence of any single biased model and improving reliability across the entire test label space.

\subsection{Interpretation and implications}

The comparative results demonstrate that the proposed framework can recommend pathology tests and handle multi-label classification effectively when built on the CC approach. By taking into account dependencies between tests, the CC-based models generate contextually suitable predictions that are closer to real diagnostic reasoning than those produced by independent classifiers. The strong performance of logistic regression with CC highlights that even relatively simple base models benefit substantially from modelling label correlations.

The robustness of the framework under natural class imbalance is particularly important from a clinical perspective. The system maintains high predictive performance without synthetic resampling, suggesting that it learns from genuine diagnostic relationships while preserving clinical realism. This characteristic is valuable in workflows where the frequency of specific tests varies significantly across patient populations and conditions.

Explainable AI analysis using SHAP adds an additional layer of validation to the clinical relevance of the model. By identifying key symptoms that drive specific test recommendations and showing that these relationships align with established medical knowledge, the explanations increase clinician trust and support the integration of the system into decision-support workflows. Overall, the combination of classifier chains, ensemble learning, and SHAP-based interpretability indicates that the proposed framework can provide reliable, transparent, and clinically usable support for early-stage diagnostic decision-making.

\section{Conclusion}

This study proposed a pathological test recommendation framework based on classifier chains and ensemble learning to enhance prediction accuracy by explicitly modelling dependencies among diagnostic tests. By framing pathology test selection as a multi-label classification problem, the approach goes beyond traditional single-label disease prediction and captures the natural co-occurrence of tests observed in clinical decision-making. The empirical results confirm that classifier-chain–based models consistently outperform independent binary classifiers across Hamming loss, accuracy, and F1-score, demonstrating that modelling label correlations yields more accurate and contextually appropriate recommendations. Explainable AI analysis using SHAP further shows that the model’s decision logic aligns with established pathology reasoning, which enhances interpretability and supports clinician trust in the recommendations.

Despite these promising results, several limitations remain. The dataset was constructed with support from physicians in a limited number of institutions, which may constrain the generalisability of the findings to other regions or healthcare settings. In addition, the current framework is based on classical machine learning models without exploring more complex deep multi-label architectures or graph-based dependency structures. The evaluation is retrospective and does not yet include prospective clinical validation or formal user studies with clinicians.

Future work will address these limitations by expanding the dataset to multi-site and more diverse populations, enabling more robust assessment of generalisability. Further research will investigate advanced dependency models, such as graph-based approaches and deep multi-label networks, to capture higher-order relationships among tests. Additional explainability techniques and structured clinician feedback studies will be used to deepen understanding of how the system is perceived in practice and to refine its interface and outputs. With these developments, the proposed pathological test recommendation framework has the potential to evolve into a deployable AI-based clinical decision-support tool, providing value to patients and physicians through more timely, accurate, and transparent data-driven medical assessments.
\section*{FUNDING INFORMATION}
Authors state no funding involved.
\section*{AUTHOR CONTRIBUTIONS STATEMENT}
Conceptualization: Abu Rafe Md Jamil, Nayan Malakar; Methodology: Abu Rafe Md Jamil, Nayan Malakar; Software and Formal Analysis: Abu Rafe Md Jamil; Investigation and Data Curation:Abu Rafe Md Jamil, Nayan Malakar; Writing—original draft preparation: Nayan Malakar; Writing—review and editing: Abu Rafe Md Jamil, Nayan Malakar; Visualization: Nayan Malakar; Supervision: Abu Rafe Md Jamil. All authors have read and approved the final manuscript.
\section*{CONFLICT OF INTEREST STATEMENT}
The authors declare that they have no known competing financial interests or personal relationships that could have appeared to influence the work reported in this paper. Authors state no conflict of interest.

\section*{DATA AVAILABILITY}
\label{}
The dataset generated and analysed during the current study is not publicly available due to institutional and privacy constraints but is available from the corresponding author on reasonable request.

\footnotesize
\itemsep 0pt 
\bibliographystyle{IEEEtran}
\bibliography{bibliography}

@article{chauhan2008disease,
  title = {Disease prediction using machine learning},
  author = {Chauhan, Raj H and Naik, Daksh N and Halpati, Rinal A and Patel, Sagarkumar J and Prajapati, M},
  journal = {Clin. Rep},
  pages = {783--787},
  year = {2008}
}

@article{rahman2023predicting,
  title = {Predicting disease from several symptoms using machine learning approach},
  author = {Rahman, Md Atikur and Nipa, Tania Ahmed and Assaduzzaman, Md},
  journal = {International Research Journal of Engineering and Technology (IRJET)},
  volume = {10},
  number = {7},
  pages = {836--841},
  year = {2023}
}

@article{ramalingam2018heart,
  title = {Heart disease prediction using machine learning techniques: a survey},
  author = {Ramalingam, VV and Dandapath, Ayantan and Raja, M Karthik},
  journal = {International Journal of Engineering \& Technology},
  volume = {7},
  number = {2.8},
  pages = {684--687},
  year = {2018},
  publisher = {Science Publishing Corporation}
}

@article{beyene2018survey,
  title = {Survey on prediction and analysis the occurrence of heart disease using data mining techniques},
  author = {Beyene, Chala and Kamat, Pooja},
  journal = {International Journal of Pure and Applied Mathematics},
  volume = {118},
  number = {8},
  pages = {165--174},
  year = {2018}
}

@article{grampurohit2020disease,
  title = {Disease prediction using machine learning algorithms},
  author = {Grampurohit, Sneha and Sagarnal, Chetan},
  journal = {2020 international conference for emerging technology (INCET)},
  pages = {1--7},
  year = {2020},
  organization = {IEEE}
}

@article{gavhane2018prediction,
  title={Prediction of heart disease using machine learning},
  author={Gavhane, Aditi and Kokkula, Gouthami and Pandya, Isha and Devadkar, Kailas},
  journal={2018 second international conference on electronics, communication and aerospace technology (ICECA)},
  pages={1275--1278},
  year={2018},
  organization={IEEE}
}

@article{el2022multi,
  title={Multi-label active learning-based machine learning model for heart disease prediction},
  author={El-Hasnony, Ibrahim M and Elzeki, Omar M and Alshehri, Ali and Salem, Hanaa},
  journal={Sensors},
  volume={22},
  number={3},
  pages={1184},
  year={2022},
  publisher={MDPI}
}

@article{shahin2014data,
  title={Data mining in healthcare information systems: case studies in Northern Lebanon},
  author={Shahin, Ahmad and Moudani, Walid and Chakik, Fadi and Khalil, Mohamad},
  journal={The Third International Conference on e-Technologies and Networks for Development (ICeND2014)},
  pages={151--155},
  year={2014},
  organization={IEEE}
}

@article{kumar2021disease,
  title={Disease prediction and doctor recommendation system using machine learning approaches},
  author={Kumar, Anand and Sharma, Ganesh Kumar and Prakash, UM},
  journal={International Journal for Research in Applied Science and Engineering Technology},
  volume={9},
  pages={34--44},
  year={2021}
}

@article{mallela2021disease,
  title={Disease prediction using machine learning techniques},
  author={Mallela, Roop Chandrika and Bhavani, Reddy Lakshmi and Ankayarkanni, B},
  journal={2021 5th International Conference on Trends in Electronics and Informatics (ICOEI)},
  pages={962--966},
  year={2021},
  organization={IEEE}
}

@article{pahwa2017prediction,
  title={Prediction of heart disease using hybrid technique for selecting features},
  author={Pahwa, Kanika and Kumar, Ravinder},
  journal={2017 4th IEEE Uttar Pradesh section international conference on electrical, computer and electronics (UPCON)},
  pages={500--504},
  year={2017},
  organization={IEEE}
}

@article{marimuthu2018review,
  title={A review on heart disease prediction using machine learning and data analytics approach},
  author={Marimuthu, M and Abinaya, M and Hariesh, KS and Madhankumar, K and Pavithra, V},
  journal={International Journal of Computer Applications},
  volume={181},
  number={18},
  pages={20--25},
  year={2018}
}

@article{shah2020heart,
  title={Heart disease prediction using machine learning techniques},
  author={Shah, Devansh and Patel, Samir and Bharti, Santosh Kumar},
  journal={SN Computer Science},
  volume={1},
  number={6},
  pages={345},
  year={2020},
  publisher={Springer}
}

@article{singh2020heart,
  title={Heart disease prediction using machine learning algorithms},
  author={Singh, Archana and Kumar, Rakesh},
  journal={2020 international conference on electrical and electronics engineering (ICE3)},
  pages={452--457},
  year={2020},
  organization={IEEE}
}

@article{zhang2007mlknn,
  title={ML-KNN: A lazy learning approach to multi-label learning},
  author={Zhang, Min-Ling and Zhou, Zhi-Hua},
  journal={Pattern Recognition},
  volume={40},
  number={7},
  pages={2038--2048},
  year={2007},
  publisher={Elsevier}
}

@article{read2011classifier,
  title={Classifier chains for multi-label classification},
  author={Read, Jesse and Pfahringer, Bernhard and Holmes, Geoff and Frank, Eibe},
  journal={Machine Learning},
  volume={85},
  number={3},
  pages={333--359},
  year={2011},
  publisher={Springer}
}
\end{document}